\documentclass{article}

\usepackage{arxiv}

\usepackage[utf8]{inputenc} % allow utf-8 input
\usepackage[T1]{fontenc}    % use 8-bit T1 fonts
\usepackage{hyperref}       % hyperlinks
\usepackage{url}            % simple URL typesetting
\usepackage{booktabs}       % professional-quality tables
\usepackage{amsfonts}       % blackboard math symbols
\usepackage{nicefrac}       % compact symbols for 1/2, etc.
\usepackage{microtype}      % microtypography
\usepackage{cleveref}       % smart cross-referencing
\usepackage{lipsum}         % Can be removed after putting your text content
\usepackage{graphicx}
\usepackage{natbib}
\usepackage{doi}
\usepackage{multirow}

\title{Compressing Cross-Lingual Multi-Task Models at Qualtrics}
\author{
    Daniel Campos\footnote{This work was done while the first author was doing an internship at Qualtrics} \\
    University of Illinois \\
    Urbana-Champagne \\
    \texttt{dcampos3@illinois.edu}
    \And
    Daniel Perry \\
    Qualtrics \\
    Seattle, WA \\
    \texttt{dperry@qualtrics.com}
    \And
    Samir Joshi \\
    Qualtrics \\
    Seattle, WA \\
    \texttt{samirj@qualtrics.com}
    \And
    Yashmeet Gambhir \\
    Qualtrics \\
    Seattle, WA \\
    \texttt{yashmeetg@qualtrics.com}
    \And
    Wei Du \\
    Qualtrics \\
    Seattle, WA \\
    \texttt{weidu@qualtrics.com}
    \And
    Zhengzheng Xing \\
    Qualtrics \\
    Seattle, WA \\
    \texttt{zxing@qualtrics.com}
    \And
    Aaron Colak \\
    Qualtrics \\
    Seattle, WA \\
    \texttt{aaronrc@qualtrics.com}
}
\begin{document}
\maketitle
\begin{abstract}
Experience management is an emerging business area where organizations focus on understanding the feedback of customers and employees in order to improve their end-to-end experiences.
This results in a unique set of machine learning problems to help understand how people feel, discover issues they care about, and find which actions need to be taken on data that are different in content and distribution from traditional NLP domains.
In this paper, we present a case study of building text analysis applications that perform multiple classification tasks efficiently in 12 languages in the nascent business area of experience management.
In order to scale up modern ML methods on experience data, we leverage cross lingual and multi-task modeling techniques to consolidate our models into a single deployment to avoid overhead.
We also make use of model compression and model distillation to reduce overall inference latency and hardware cost to the level acceptable for business needs while maintaining model prediction quality.
Our findings show that multi-task modeling improves task performance for a subset of experience management tasks in both XLM-R and mBert architectures.
Among the compressed architectures we explored, we found that MiniLM achieved the best compression/performance tradeoff.
Our case study demonstrates a speedup of up to 15.61x with 2.60\% average task degradation (or 3.29x speedup with 1.71\% degradation) and estimated savings of 44\% over using the original full-size model.
These results demonstrate a successful scaling up of text classification for the challenging new area of ML for experience management.
\end{abstract}

\section{Introduction}
Experience management enables businesses and organizations to effectively adapt to actionable feedback from their customers and employees.
Understanding and managing a customer or employee experience requires analyzing a combination of survey questions, social media, and other sources of experience data together in order to derive insight.
Deriving insight requires effectively predicting the feelings, emotions, as well as the requested actions from feedback in a scalable way.
To accurately predict these insights we leverage state-of-the-art pretrained language models.

However, many of the best pretrained models require significant resources to deploy at scale. 
For example, the best performing models for tasks such as semantic similarity of sentences~\cite{wang2018glue} can have hundreds of billions of parameters~\cite{smith2022using}.
Even more pedestrian (in terms of size) models such as BERT-base~\cite{devlin-etal-2019-bert} can still be relatively expensive and latency-prone for a typical business use case, especially without specialized hardware or accelerators.
One way to both achieve high prediction accuracy and scale up is to leverage model compression.

While there is substantial literature on model compression, it can be difficult to sort through all the methods and evaluate them on a case by case basis.
Our contribution is a specific case study evaluation of model compression methods for Qualtrics models on experience management data and its unique challenges.
In this work, we are particularly interested in building efficient text classifiers, an important problem in the experience management domain. 
Indeed, unstructured data constitutes more than 80\% of the experience management data.
As such, analyzing text data across dimensions such as sentiment, emotion, actionability, effort, intent, topic, urgency, and toxicity is one of the most the foundational challenges in this emerging space. 
We share details about what worked and did not, which can benefit the industry at large as others begin to adopt model compression in their organizations for their use cases.
This is particularly timely in our current market as many companies in emerging business areas are looking to reduce costs and model compression is an effective way to reduce ML inference costs, both financially and environmentally.

\subsection{Motivating Constraints: Engineering Overhead, Cost, Latency}
Our goal in pursuing this compression and consolidation work was to reduce overall model hosting costs while preserving model quality.
Two areas we are focused on are reducing burden for engineering support of hosting our new models in production, as well as the direct cost and latency of the models themselves.

Since we deploy and support our models in a microservices framework, each model typically lives behind a specific model endpoint or service so each model has a static cost for the base capacity, and variable cost for the elastic capacity.
If we use single-task monolingual models, this results in needing to support in production a specific service per task language pair.
Similarly, for single-task NLP models, the encoder, which can account for 90+\% of the computation for classification models, must run for each task, regardless of how similar it is between tasks.

In contrast, a multi-task cross-lingual model consolidates this repetitive computation and removes the instance hosting overhead for additional languages.
For this reason, we focused on the ability to support multiple tasks per model, as well as a cross-lingual model.

In addition, by developing smaller models, we hope to achieve reduced latency for runtime needs while also reducing costs by providing flexibility to deploy on less costly hardware.

\subsection{The Tension Between Model Consolidation and Compression}
There is an interesting tension that arises as we both combine multiple models into a single multi-task cross-lingual model and also reduce the size and capacity of that model.
While prior work has also looked at these different facets of model consolidation and compression in isolation \cite{wang2020minilm, wang2020minilmv2, mukherjee2021xtremedistiltransformers, jiao2021lightmbert, sanh2019distilbert, jiao2020tinybert, de2020optimal,yang2019model}, in this work we investigate how these approaches work together to consolidate and compress a model, and how that impacts model performance on the target tasks.

We are unable to analyze this tension for all NLP tasks in general, but here we present evidence towards understanding the tradeoffs for specific cases, relevant to work at our company.
These results can inform future theoretical work as well as more practical application at other organizations.
\section{Cross-Lingual Multi-Task (XLMT) Model Compression Methods}
As described above, we are motivated to both consolidate task and language support into a single cross-lingual multi-task (XLMT) model and at the same time pursue a compressed version of that model to reduce capacity and make the model faster and less expensive to run.

\subsection{Cross-lingual Modeling}
There has been a strong movement towards multi-lingual and cross-lingual models.
One of the first multi-lingual BERT models was ``multi-lingual BERT'' (mBert), from \cite{devlin-etal-2019-bert}, which extended ``monolingual BERT'' by training across a dataset with multiple languages represented.
Cross-lingual modeling (XLM), presented in \cite{conneau2019cross}, further improved over multi-lingual modeling by introducing additional cross-lingual pretraining tasks, and 
XLM-Roberta (XLM-R) \cite{XLMR} developed a strong cross-lingual model using techniques from Roberta \cite{roberta} and showed better performance beyond previous multi-lingual and cross-lingual models.

In this work we show results using both the mBert and XLM-R pretrained models on which we build our task-specific classifiers.
In the original paper \cite{XLMR} the authors showed a decrease in model performance as more and more languages were introduced.
We explore the effect of training on monolingual vs cross-lingual settings, and how it impacts our combined model performance.

\subsection{Multi-task Learning for NLP}
Multi-task learning (MTL) can not only merge tasks into a single model but also improve task performance by sharing common layers.
For instance,~\cite{lin2018multi} proposed an architecture that shares the same character embedding layer showing effective results for low-resource settings.
Other types of MTL include hierarchical architectures, such as~\cite{vijayaraghavan2017twitter} where separate tasks are learned and then combined using a final attenuation layer and \cite{he2019interactive} where the first task output feeds into a second task in sequence.

In this work we explore how combining multiple tasks into a single cross-lingual model impacts performance on each of those tasks individually.
Our approach leverages a common base model with multiple task heads.  
The multi-task multiclass classification loss function we use consists of a simple sum of cross-entropy losses,
\begin{equation}
L_{\mbox{MT}} = \frac{1}{N} \sum_{t=1}^T \sum_{i=1}^{N^t} \left[ -\sum_{c \in C_{t}} \left( \ell_{i,c}^t \log p_{i, c}^t \right) \right],
\end{equation}
where $N = \sum_{t=1}^T {N^t}$ is the total number of data points from $T$ tasks and $N^t$ is the number of data points for the $t$-th task. $C_{t}$ is the number of classes for task $t$. $\ell_{i,c}^t$ is either $0$ or $1$, indicating whether class label $c$ is the correct classification of the $i$-th data point from the $t$-th task, and $ p_{i,c}^t$ are the corresponding predicted probabilities.  

\subsection{Model Compression}
\subsubsection{Knowledge Distillation}
Knowledge distillation (KD) popularized by \cite{hinton2015distilling} and aims to create smaller models which approximate the performance of the larger models by teaching the smaller model (student model) to emulate the larger model (teacher model). 

The original approach used the final layer logit-based knowledge distillation, where the concept is to minimize the distance (i.e., KL divergence loss function) of logit output (final layer) between teacher and student models. 
Later work, including many applications in NLP, introduced variations on this idea, including \cite{sanh2019distilbert} which applied a combined loss, including masked language modeling loss, cosine distance loss, and KL divergence loss to reduce BERT model size.
More generally, we can also align the intermediate features between the teacher and the student models rather than just the final layer, such as \cite{jiao2020tinybert} which used many intermediate layers for distillation.
MiniLM was introduced in~\cite{wang2020minilm} using self-attention distribution transfer and self-attention value-relation transfer to achieve competitive performance in both monolingual and multilingual models. \\

In this work, we have primarily investigated distilling using the task specific logits produced by the final layer.
Exploring additional intermediate representation distillation is left to future work to potentially improve performance in the smallest models we tested.
Focusing on the last layer results in the following modified loss:
\begin{eqnarray}
L_{\mbox{MT-KD}} &=& \nonumber \\
 \frac{1}{N} \sum_{t=1}^T \sum_{i=1}^{N^t} &&\left[ - \left( \sum_{c \in C_{t}} \ell_{i,c}^t \log p_{i,c}^t \right) \right. \nonumber \\
&& \left. + \alpha F^2 \left( \sum_{c \in C_{t}} \hat{q}_{i,c}^t \log \frac{\hat{q}_{i,c}^t}{\hat{p}_{i,c}^t} \right)  \right],
\end{eqnarray}
where $q_{i,c}^t$ is the teacher model prediction of the $i$-th data point from the $t$-th task, $\hat{q}_{i,c}^t = \frac{\exp(q_{i,c}^t/F)}{ \sum_j \exp(q_{j,c}^t/F)}$ is the temperature modified teacher prediction, $\hat{p}_{i,c}^t = \frac{\exp(p_{i,c}^t/F)}{ \sum_j \exp(p_{j,c}^t/F)}$ is the temperature modified student prediction, $F$ is the temperature parameter \cite{hinton2015distilling}, and $\alpha$ is the teacher coefficient term controlling the relative impact of distillation to the label loss.

\subsubsection{Structural Pruning}
In~\cite{LeCun1989OptimalBD} the author introduced a notion that neural networks can be compressed by removing entire sections without major impact to accuracy. 
Structural pruning compress networks by removing entire structural components like attention heads, neurons, and even transformer layers, and leverage KD to limit model degradation.
While most previous work in compression has focused on monolingual models, there is also a growing body of work around multilingual and cross-lingual model compression \cite{jiao2021lightmbert,mukherjee2020xtremedistil,mukherjee2021xtremedistiltransformers,wang2020minilm,wang2020minilmv2}.
We focus on two specific compressed architectures, MiniLM \cite{wang2020minilmv2} and XtremeDistil \cite{mukherjee2021xtremedistiltransformers} and compare them in our use case.
Ultimately we found MiniLM to be the most effective at learning our specific set of tasks.

\subsubsection{Quantization}
Quantization enables reduction in model size and memory footprint while also potentially increasing inference speed.
Here we consider integer quantization, in which the precision is reduced from 32-bit floating point to 8-bit integer.
Quantization can be done during training, known as quantization aware training (QAT), to minimize degradation, or after training, known as post training quantization (PTQ), to compress an already trained model.
\cite{zafrir2019} shows that by leveraging QAT, their "Q8Bert" quantized model was able to match the performance of the base BERT model on various NLP tasks.

In this work we explore combining quantization via QAT with structural pruning to further reduce the model size while maintaining good model performance.

\section{Experimental Results}
Our core set of results are developed around a multi-task cross-lingual model developed internally at Qualtrics to help develop understanding around customer feedback.
The model handles three separate but related multiclass classification tasks on text input, we refer to these tasks throughout this paper as Task-1, Task-2, and Task-3.
They refer to three text classification tasks our group actively uses and develops, with similarities to models such as sentiment or toxicity prediction \cite{he2019interactive}.
Each of these task is implemented as a sequence classification task where the input is direct customer feedback. 
Task-1 is a multi class classification with 6 labels, 
Task-2 is a multi-class classification with 4 labels, and
Task-3 is a multi-class, multi-label sequence classification with 9 classes and each class has independent binary labels. \\

In our experiments our focus is on exploring the relationship between knowledge distillation, multi-task modeling, quantization and multilingualism. 
We do not seek to provide a complete understanding of how each axis impacts the outcomes but instead seek to find the optimal way to optimize the performance of pruned and quantized model by exploring the impact of multi-lingual fine tuning, variations in knowledge distillation, and task specific teachers.

\subsection{Dataset and Compressed Architecture Selection}
Our dataset consists of a collection of internal customer experience data across multiple industries.
The data has been fully anonymized and aggregated, and is used with permission.
This process protects customer data privacy and ensures data from any specific industry or company is not over-represented or identifiable.
The resulting text dataset consists of 257k text documents across 16 languages labeled for Task-1, and 127k text documents across 12 languages labeled for Task-2 and Task-3.
A description of the task types, number of labels, and label can be seen in Table~\ref{table_dataset_breakdown}.
This experimental data is similar in nature to the production datasets used in our production modeling system.

\begin{table}[h]
    \centering
    \begin{center}
    \begin{tabular}{ l|ll|ll|l }
        \hline
        \multicolumn{2}{c}{\textbf{Task-1}} & \multicolumn{2}{||c||}{\textbf{Task-2}} & \multicolumn{2}{c}{\textbf{Task-3}} \\
        \multicolumn{2}{c}{(Multi-class)} & \multicolumn{2}{||c||}{(Multi-class)} & \multicolumn{2}{c}{(Multi-label)} \\
        \hline
        \multicolumn{2}{c}{\# Samples} & \multicolumn{2}{||c||}{\# Samples} & \multicolumn{2}{c}{\# Samples} \\
        \hline
        \multicolumn{1}{c}{T1-L1} & \multicolumn{1}{|c}{93738} & \multicolumn{1}{||c|}{T2-L1} & \multicolumn{1}{c||}{83545} & \multicolumn{1}{c|}{T3-L1} & \multicolumn{1}{c}{33463} \\
        \multicolumn{1}{c}{T1-L2} & \multicolumn{1}{|c}{70218} & \multicolumn{1}{||c|}{T2-L2} & \multicolumn{1}{c||}{36018} & \multicolumn{1}{c|}{T3-L2} & \multicolumn{1}{c}{21562} \\
        \multicolumn{1}{c}{T1-L3} & \multicolumn{1}{|c}{38786} & \multicolumn{1}{||c|}{T2-L3} & \multicolumn{1}{c||}{4317} & \multicolumn{1}{c|}{T3-L3} & \multicolumn{1}{c}{22556} \\
        \multicolumn{1}{c}{T1-L4} & \multicolumn{1}{|c}{26359} & \multicolumn{1}{||c|}{T2-L4} & \multicolumn{1}{c||}{3198} & \multicolumn{1}{c|}{T3-L4} & \multicolumn{1}{c}{1090} \\
        \multicolumn{1}{c}{T1-L5} & \multicolumn{1}{|c}{18792} & \multicolumn{1}{||c|}{} & \multicolumn{1}{c||}{} & \multicolumn{1}{c|}{T3-L5} & \multicolumn{1}{c}{7525} \\
        \multicolumn{1}{c}{T1-L6} & \multicolumn{1}{|c}{9837} & \multicolumn{1}{||c|}{} & \multicolumn{1}{c||}{} & \multicolumn{1}{c|}{T3-L6} & \multicolumn{1}{c}{44485} \\
        \multicolumn{1}{c}{} & \multicolumn{1}{|c}{} & \multicolumn{1}{||c|}{} & \multicolumn{1}{c||}{} & \multicolumn{1}{c|}{T3-L7} & \multicolumn{1}{c}{11341} \\
        \multicolumn{1}{c}{} & \multicolumn{1}{|c}{} & \multicolumn{1}{||c|}{} & \multicolumn{1}{c||}{} & \multicolumn{1}{c|}{T3-L8} & \multicolumn{1}{c}{2518} \\
        \multicolumn{1}{c}{} & \multicolumn{1}{|c}{} & \multicolumn{1}{||c|}{} & \multicolumn{1}{c||}{} & \multicolumn{1}{c|}{T3-L9} & \multicolumn{1}{c}{1951} \\
        \hline
    \end{tabular}
    \end{center}
    \caption{Breakdown of task label distribution. Task labels are listed as T\#-L\#, where T1-L1 represents Label 1 for Task-1.}
    \label{table_dataset_breakdown}
\end{table}
For modeling we primarily use PyTorch \cite{NEURIPS2019_9015} and the Transformers library \cite{wolf2019huggingface}.
For model quantization, we made use of the SparseML library \cite{pmlr-v119-kurtz20a, singh2020woodfisher}.

Instead of developing our own target architecture, we leverage existing cross-lingual models from the literature as a first approach to model compression.
After a review of the literature, we settled on experimentation around two cross-lingual models, XtremeDistil \cite{mukherjee2020xtremedistil}, \cite{mukherjee2021xtremedistiltransformers} and MiniLM \cite{wang2020minilm}, \cite{wang2020minilmv2}.
We summarize the characteristics of the architectures evaluated in Table~\ref{table_archs}, where the smallest model considered was 6 layers and 22M parameters.

\begin{table}[h]
\centering
\begin{center}
\begin{tabular}{cccc}
    \hline
    \textbf{Name} & \textbf{\#Layer} & \textbf{\#Param} & \textbf{Size}\\
    \hline
    XLM-R(XLM-R Base) & 12 & 85M & 1.12GB \\
    XtremeDistil & 6 & 22M & 91MB \\
    MiniLM-L12(mMiniLMv2) & 12 & 41M & 236MB \\
    MiniLM-L6(mMiniLMv2) & 6 & 22M & 215MB \\
    \hline
\end{tabular}
\end{center}
\caption{Description of model architectures evaluated. \#Params refers to the number of transformer parameters. }
\label{table_archs}
\end{table}

To further narrow to a single cross-lingual model, we performed an experiment using a subset of our datasets that covered 11 languages and evaluated how well the models perform in two settings: with a distillation teacher and without a teacher.
The subset contained 57k responses labeled for Task-1 and 20k labeled for Task-2 and Task-3.

\begin{table}[h]
\centering
\begin{tabular}{cccc} 
    \hline
    \textbf{Method} & \textbf{Task-1} & \textbf{Task-2} & \textbf{Task-3} \\
    \hline
    XLM-R & 83.32 & 80.81 & 39.41 \\ \hline
    Xtremedistil \hspace{1cm} (no teacher) & 67.69 & 69.24 & 31.23 \\
    Xtremedistil (with teacher) & 67.82 & 70.99 & 28.93 \\
    MiniLM-L12 \hspace{1cm} (no teacher) & 80.79 & 77.55 & 35.99 \\
    MiniLM-L12 \hspace{1cm} (with teacher) & \textbf{81.43} & \textbf{78.44} & \textbf{36.54} \\
    \hline
\end{tabular}
\caption{Results on each task for each model architecture, reported in Macro-F1. All models were trained for 2 epochs and reported results are the per-task macro F1 scores.}
\label{table_arch_search}
\end{table}

This experiment, as shown in Table~\ref{table_arch_search}, indicated that MiniLM (and its variants) would be easier to train and perform model distillation in our setting.
Due to the above results, for compressed models we targeted the MiniLM-L12 architecture.
Our definition of performing better, worse, or the same was based on the 95th percentile confidence interval of a random sample of 5 models trained from different random seeds.
If we observe differences greater than these intervals we consider them significant; otherwise we consider the result to be the same.

\subsection{Cross-Lingual Model Results}
Our goal in developing a cross-lingual model is to reduce the overhead of hosting multiple monolingual models.
However, the single cross-lingual model should perform at least on par with the monolingual model.
To test this assumption, we trained a single cross-lingual model and tested it across all languages.
We then trained 12 separate monolingual models, starting from the publicly available XLMR-base pretrained weights (to avoid confounding factors from alternative monolingual base models).
We then evaluated these monolingual models against the same cross-lingual evaluation dataset as a benchmark.
A summary of results is shown in Table~\ref{table:xlm}, where we report results for \emph{fr}, \emph{en}, \emph{de}, and \emph{ja} languages. 
We also evaluated 8 other languages and observed the same overall relative results. 
The best monolingual Task-1 result overall was $73.39$ (\emph{en}) and worst was $14.52$ (\emph{pl}), the corresponding cross-lingual reaching $79.12$ (\emph{pl}). 
The best cross-lingual result was $91.65$ (\emph{en}) and the worst was $71.05$ (\emph{ko}) with the corresponding monolingual result dipping to $48.84$ (\emph{ko}).
We observe in every language we examined the cross-lingual model does better than the monolingual model, strongly supporting a move to cross-lingual modeling for our tasks.

\begin{table}[h]
\centering
\begin{tabular}{ccccc}
    \hline
    \textbf{Train Lang} & \textbf{Eval Lang} &  \textbf{Task-1} & \textbf{Task-2} & \textbf{Task-3} \\
    \hline
    all & \multirow{2}{*}{fr} & \textbf{87.69} & \textbf{82.43} & \textbf{36.16} \\
    fr & & 68.91 & 74.04 & 26.24 \\
    \hline
    all & \multirow{2}{*}{en} & \textbf{91.65} & \textbf{79.67} & \textbf{40.47} \\
    en & & 73.39 & 77.2 & 33.69 \\
    \hline
    all & \multirow{2}{*}{de} & \textbf{86.07} & \textbf{77.74} & \textbf{34.71} \\
    de & & 68.71 & 70.65 & 23.52 \\
    \hline
    all & \multirow{2}{*}{ja} & \textbf{80.3} & \textbf{70.71} & \textbf{32.2} \\
    ja & &  56.22 & 64.21 & 15.9 \\
    \hline
\end{tabular}
\caption{Cross-lingual model comparison with monolingual models, evaluated on the same target language.  Across all languages and tasks we evaluated, we observed the cross-lingual models to outperform monolingual models.}
\label{table:xlm}
\end{table}

\subsection{Cross-Lingual Multi-Task (XLMT) Model Results}
We are also interested in combining multiple tasks into a single model to reduce the engineering overhead of hosting our model.
To evaluate whether our model maintained similar performance to single-task models, we evaluated the combined XLMT model in comparison to the single-task models for both XLM-R and mBert pretrained models.
The experimental results in Table~\ref{table:tasks} show that the XLMT model performed similarly to, if not better than, the single-task model on Task-1 and Task-2 prediction.
For Task-3 we observed some significant degradation in the task performance.

To further confirm these results, we performed a similar analysis using another multilingual model, mBert.
Using mBert, we again observed some modest gains for the first two tasks and then significant degradation for the third task.

These results indicate our current multi-task architecture does benefit two of the three tasks. 
However, for final deployment it will be important to consider moving our third task into a separate model or develop alternative multi-task architectures to reduce the performance gap.

\begin{table}[h]
\centering
\begin{tabular}{c|ccc}
    \hline
    \textbf{Model} & \textbf{Train Method} & \textbf{Eval Task} & \textbf{F1} \\
    \hline
    \multirow{6}{*}{XLM-R} & \multirow{3}{*}{multi-task}& Task-1 & 82.23 \\
     &  & Task-2 & 76.03 \\
     &  & Task-3 & 38.32 \\
     \cline{2-4}
     & \multirow{3}{*}{single-task} & Task-1 & 81.12 \\
     &  & Task-2 & 74.67 \\
     &  & Task-3 & 51.27 \\
    \hline
    \multirow{6}{*}{mBert} &  \multirow{3}{*}{multi-task} & Task-1 & 78.88 \\
     &  & Task-2 & 75.27 \\
     &  & Task-3 & 35.88 \\
     \cline{2-4}
     &  \multirow{3}{*}{single-task} & Task-1 & 78.63 \\
     &  & Task-2 & 74.31 \\
     &  & Task-3 & 51.12 \\
    \hline
\end{tabular}
\caption{Task-specific results for cross-lingual single-task models and multi-task models. Macro-F1 results are reported on the full evaluation set, consisting of all languages (16 for Task-1, 12 for Task-2/3).}
\label{table:tasks}
\end{table}

\subsection{Compressed XLMT Model Results}
In developing the XLMT model, engineering overhead was reduced from $16 \times 1 + 12 \times 2 = 40$ individual models to a single cross-lingual multi-task models, or two based on the outcomes of Task-3 above.
However, given the size of the XLM-Roberta model, the hosting costs associated with serving inference, specifically given the need for GPU instances to generate predictions at low latency, remained high.
To reduce this base cost and reduce the latency of this model, we focused on compressing the model itself.
As mentioned earlier this compressing of the model, reducing its overall capacity, is in tension with the goals of maintaining performance of the combined XLMT model.

Our results in Table~\ref{table:pruning} show that simply performing structured layer pruning on the model resulted in some degradation of task performance.  
For Task-1 with MiniLM-12 architecture, the larger of the smaller architectures considered, we see about $1.6$\% relative degradation. MiniLM-6 shows 3.9\% degradation, while XtremeDistil shows over $20$\% degradation.
This same pattern holds for the Task-2, and for Task-3 we see even less degradation for MiniLM-12.

These results strongly favor MiniLM-12 and MiniLM-6 for compressing our specific use case.

\begin{table}[h]
\centering
\begin{tabular}{cccc}
    \hline
    \textbf{Model} &  \textbf{Task-1} & \textbf{Task-2} & \textbf{Task-3} \\
    \hline
    XLM-R & 82.23 & 76.80 & 35.90 \\
    MiniLM-L12 & 80.85 & 75.86 & 35.09 \\  % average degradation is 1.71%
    MiniLM-L6 & 78.97 & 72.42 & 35.34 \\
    XtremeDistil & 61.83 & 61.59 & 24.00 \\
    \hline
\end{tabular}
\caption{Results comparing the original MiniLM and XtremeDistil models with the full-size XLM-R model across Task-1, Task-2, and Task-3 macro-F1 scores.}
\label{table:pruning}
\end{table}

\subsection{Distilled XLMT Model Results}
To address the degradation resulting from structured layer pruning we incorporated some model distillation using the final layers of the full-size and compressed models.
We explored using a single multi-task teacher and task-specific teachers, as well as using a single cross-lingual teacher and language-specific teachers.
However we ultimately use cross-lingual task-specific teachers because the performance of Task-3 as a single task model outperformed the multi-task model, as shown in Table~\ref{table:tasks} and cross-lingual models consistently out-performed language-specific models as shown in Table~\ref{table:xlm}.
To provide additional model compression after enabling distillation, we trained the model with QAT in order to further reduce model complexity.
To evaluate model speedup, each model was run for sequences of length 128 with batch size 32 on 1 Nvidia T4 GPU leveraging TensorRT \footnote{\url{https://developer.nvidia.com/tensorrt}}.
Speedup was measured relative to the baseline model, XLM-R (fp32).

\begin{table}[h]
\centering
\begin{tabular}{ccccc}
    \hline
    \textbf{Model} & \textbf{Speedup} &  \textbf{Task-1} & \textbf{Task-2} & \textbf{Task-3} \\
    \hline
    XLM-R \hspace{1cm} (fp32) & x1 & 82.23 & 76.80 & 35.90 \\
    XLM-R \hspace{1cm} (int8 quantized) & x3.64 & 81.09 & 73.60 & 35.80 \\
    \hline
    MiniLM-L12 \hspace{1cm} (fp32) & x3.29 & 79.42 & 75.1 & 35.36 \\		% average degradation is 2.37%
    MiniLM-L12 \hspace{1cm} (int8 quantized) & x8.11 & 79.29 & 73.69 & 35.71 \\  % average degradation is 2.71%
    \hline
    MiniLM-L6 \hspace{1cm} (int8 quantized) & x15.61 & 79.05 & 73.90 & 35.84 \\  % average degradation is 2.60%
    \hline
    \hline
    MiniLM-L12-mBert \hspace{1cm} (int8 quantized) & x8.11 & 79.05 & 73.48 & 35.57 \\
    \hline

\end{tabular}
\caption{
Results on model distillation and quantization aware training. 
Task-1, Task-2, and Task-3 results are reported in macro-F1 scores.  
XLM-R models were used as the teacher in all results, except for MiniLM-L12-mBert which used mBert teachers.
}
\label{table:distill}
\end{table}

The two best models after distillation were the quantized MiniLM-L6 model with $2.60$\% average relative task degradation and non-quantized MiniLM-L12 with $2.37$\% average relative task degradation.
We found that quantized MiniLM-L6 was able to improve more from distillation than MiniLM-L12.
While we are still investigating full cause our current hypothesis is that the smaller model provides some regularization against overfitting versus the larger model.
In terms of speedup our quantized MiniLM-L6 model provided the most speedup at 15.61x speedup over the baseline.
In our final assessment we found that using task-specific model distillation on the MiniLM-L6 model with quantization provided a strong result in model size while maintaining model performance, as shown in Table~\ref{table:distill}.
However, in considering the best model overall the MiniLM-L12 in Table~\ref{table:pruning} provided the least overall degradation of 1.71\% and a modest speedup of 3.29x.

\section{Business Impact}
An implementation of these compression techniques and choices was developed in our production training system.
At Qualtrics, this system has generated significant business impact across customers and new features, financial, environmental, and operational cost, system flexibility and robustness.

\subsection{Feature impact}
Given the speedup the compressed and multi-task models provide, we observe significant increases in throughput across use cases, enabling us to serve more customers, and enabling new features for customers that were previously too compute intensive.
As an example, this speedup enables ML pipelines that run 3-4 models serially in the same time window as a single model.
Additionally, the flexibility of cross-lingual models enables us to serve more customers in more languages, without large training sets or comprehensive language specialization.

\subsection{Financial impact}
Conservatively, we estimate approximately 44\% savings in terms of hardware cost from the developments in compression of the multi-task cross-lingual model, in comparison to an uncompressed system at similar latency and throughput.
In addition, by combining multiple tasks that are invoked with similar load into a single model, we achieve a fraction of the total inference cost.

This savings is driven by several factors: reducing base instance capacity; reducing the amount of dynamic scaling; and allowing for deployment of lower cost hardware.
We note that the savings is limited by the persistent cost of the base capacity, even when reduced, which creates a floor for cost savings even with models that are multiple times faster.

Currently, we deploy our models on a public cloud with a cost of approximately \$80K/month/model.
The compression technique used in this paper reduces cost by 44\%, resulting in \$35K savings/month for a single model compared to the current model in production for the same tasks.
As we push our compression framework to various other NLP systems currently developed by this group, it can result in a potential yearly cost savings in the single digit millions of dollars.
Considering current macro-economic conditions when there is industry wide need for financial costs reduction, the significant savings can strengthen the fundamentals of a typical SaaS company like Qualtrics. 

\subsection{Ethical impact}
We are pleased that our efforts to compress models will support environmental sustainability.
By reducing the amount of power and resource needed to run the same inference, we anticipate meaningful impact on environmental footprint, though we are unable to quantify it concretely at this time.

\subsection{Operational impact \& robustness}
While MiniLM-L6 provided better speedup, business needs required the lower degradation provided by MiniLM-L12 for the first set of models.
By leveraging the compressed XLMT model, we enable additional flexibility in production deployment scenarios, including different instance types (CPU, GPU, custom accelerators) that enable us to serve high throughput inference while balancing cost.
This was not previously viable with larger models, which required GPUs in order to serve high throughput inference.
By enabling this flexibility, we also improve system robustness, as the models are robust to instance unavailability or instance downtime for any type of instance.

Additionally, savings from moving to a single multi-task model results in overall reduced workload for our engineering teams, removing per-task deployments and deployment pipelines and lowering the barriers to new tasks and new language support.
Specifically, multi-task and cross-lingual modeling reduces the number of models for these tasks from 36 potential models (12 languages, 3 tasks) to a single model, reducing operational cost from 6-7 on-call/operations engineers to 1.
Compression additionally reduces this cost by lowering latency and increasing throughput, reducing the operational cost of mitigating latency spikes and scaling issues.

\section{Conclusion}
We have presented a case study into the Qualtrics approach for leveraging cross-lingual and multi-task modeling techniques in combination with model distillation and quantization techniques, in order to develop models that can handle traffic volumes at scale.
The results show that for our multi-class classification tasks, these methods can be combined effectively to reduce deployment overhead and maintenance and achieve up to 15.61x speedup with 2.60\% average degradation.
Our results explore boundary cases where compression works well, and where it can degrade past business requirements; where combining up to 12 languages works well; and where combining tasks works well, and where it does not.
These approaches have been necessary for us to scale up our unique text classification problems in the growing field of experience management.
We anticipate these results will help to guide other groups hoping to reduce model inference costs, as well as contribute to future theoretical work around the tradeoff between model compression and model consolidation.
Looking forward, we plan to apply these methods to more complex sequence labeling tasks and to explore additional methods such as model sparsity and neural architecture search to see if even faster models can be developed with acceptable levels of model performance.

\bibliographystyle{unsrtnat}
\bibliography{paper_main}  %%% Uncomment this line and comment out the ``thebibliography'' section below to use the external .bib file (using bibtex) .





% \end{thebibliography}

\end{document}